\documentclass[]{article}
\usepackage{graphicx}
 \usepackage{wasysym}
 \usepackage{listings}
 \usepackage{fullpage}
\begin{document}
\title{A  Two-Systems  Perspective for Computational Thinking}
%
%
\author{Arvind W Kiwelekar $^1$,
Swanand Navandar$^1$, Dharmendra K. Yadav$^2$}

\thanks{$^1$Department of Computer Engineering,\\Dr Babasaheb Ambedkar Technological University \\ Lonere-Raigad 402103 India \\
{\{awk,snavandar\}@dbatu.ac.in} \\
$^2$Department of Computer Engineering,\\
Motilal Nehru National Institute of Technology, Allahabad \\ 
Prayagraj - 211004, India \\
{dky@mnit.ac.in} 
}

This is a pre-print of the following paper: Arvind W. Kiwelekar,  Swanand Navandar and Dharmendra K Yadav, {\em A Two systems Perspective for Computational Thinking,} accepted version for 12th International Conference on Intelligent Human Interaction (IHCI 2020) held from 24th to 26th November 2020 at Exco-Daegu South Korea.

{\bf Cite this chapter as:}
Arvind W Kiwelekar, Swanand Navandar,  and Dharmendar K Yadav, {\em A Two systems Perspective for Computational Thinking},    in the proceedings of  12th International Conference on Intelligent Human Interaction (IHCI 2020) Exco-Daegu South Korea November 2020 Publisher:  Springer.

\newpage

\maketitle              
\begin{abstract}
Computational Thinking (CT) has emerged as one of the vital thinking skills in recent times, especially for Science, Technology, Engineering and Management (STEM) graduates. Educators are in search of underlying cognitive models against which CT can be analyzed and evaluated.  This paper suggests adopting Kahneman's two-systems model as a framework to understand computational thought process.  Kahneman's two-systems model postulates that human thinking happens at two levels, i.e. fast and slow thinking. This paper illustrates through examples that CT activities can be represented and analyzed using Kahneman's two-systems model. The potential benefits of adopting Kahneman's two-systems perspective are that it helps us to fix the biases that cause errors in our reasoning. Further,  it also provides a set of heuristics to speed up reasoning activities.
\end{abstract}

\section{Introduction}

Computational  Thinking is emerging as a  generic skill for everyone, whether someone is a Computer Programmer, Data Scientist,  Biologist, Physician, Lawyer, or an ordinary human being. The pioneers who coined CT \cite{wing2006computational} often equate it to general skills like reading, writing, and speaking to highlight the broader applicability of CT.  Because  CT being the most relevant skill to learn, educators have been designing specialized curricula \cite{iyer2019teaching} to impart this skill from the kindergarten level.   Also, educators \cite{denning2019computational} are differentiating the {\em CT for Beginners} from the {\em CT for Professionals} so that it can be applied by working professionals to address their domain-specific problems.

In its simplest sense, Computational Thinking is a specialized type of human thinking required to solve problems through computers.  CT  being a human thought process, it gets influenced by human psychological traits such as attention, memory, impressions, feelings, opinions, biases and heuristics.  This paper highlights the necessity of  a rich framework grounded in Psychological theories to analyze the influences of these traits on CT. 

Hence this position paper suggests  adopting one such theory  i.e. Kahneman's two-systems model of thinking, from the field of Psychology, to analyze the human aspects involved in computational thinking.  The article defines Computational Thinking and elements of Kahneman's two systems model of thinking in Section 2, 3, and 4. Section 5 identifies CT activities and map them on Kahneman's two systems model of thinking.  Section 6 relates our proposal with the existing applications of Kahneman's model of thinking in Computer Science. The paper concludes with directions for future work.

\section{Computational Thinking (CT)} 
In the seminal paper on {\em Computational Thinking}, J M Wing \cite{wing2006computational}  clearly explains the breadth of the CT as a thought process.  This broad definition of CT includes a set of skills such as solving a problem using computers, designing and evaluating a complex system, and understanding human behaviour.  

For computer programmers, CT is a way of representing a real-life problem and solving it with the help of computers.  For example, writing a program to find an optimal path to travel from a source to a destination.   

For software engineers, CT refers to designing and evaluating a complex information processing system such as an online railway or an airline reservation system.

For computer scientists, CT refers to getting insights about human behaviour by answering questions such as (i) {\em what are the limitations and power of computation?}, (ii) {\em what does it mean by intelligence?}, (iii) {\em what motivates us as a human being to perform or not to perform a specific action?}. 

This broad coverage of topics included under computational thinking highlights that computational thinking is beyond mere computer programming or coding. 

Further, Aho \cite{aho2012computation} brings out the differences between the terms {\em Computation} and {\em Computational Thinking}. He recommends that the term {\em Computation} shall be restricted to denote those tasks for which the semantics can be described through a formal mathematical model of computation(e.g., Finite Automata, Pi-Calculus, Turing Machine).  The tasks for which  no such appropriate models exist, it is necessary to invent such formal models.

Computational thinking being a complex thought process, the paper proposes to analyze it through the cognitive model of thinking propagated by Psychologist, Economist, and Nobel Laureate Prof. Daniel Kahneman. 
Though the Kahneman's model of thinking is not a formal model useful to describe exact semantics of CT activities, the cognitive model helps us to fix the errors in our reasoning and to sharpen our thought process.   
\begin{figure}[t]
    \centering
    \includegraphics[scale=0.37]{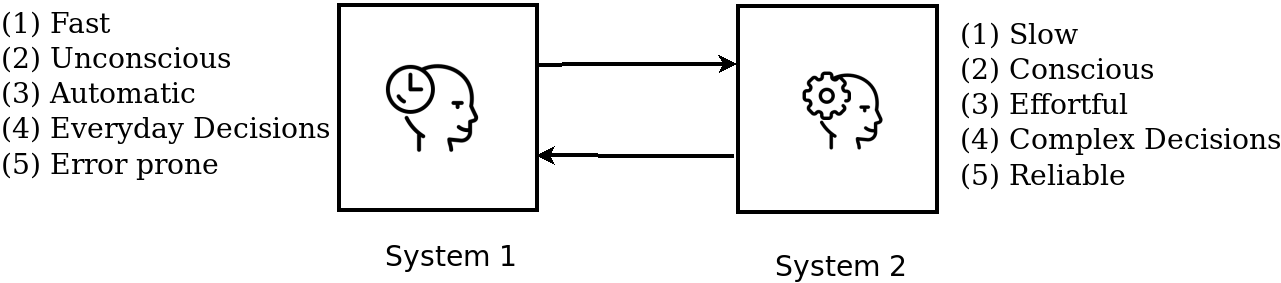}
    \caption{Kahneman's Two-Systems Model}
    \label{2systems}
\end{figure}

\section{Kahneman's Two-Systems Model of Thinking} 

The human cognitive processes, such as judgement and decision making, are complex and intricate. To understand these processes in a better and simplified way, many psychologists have proposed that human thinking operates at two different levels \cite{evans2013dual}. First one is a fast, intuitive, and effortless way of thinking requiring less or no attention. The second one is a slow, intentional, and effortful way of thinking, often requiring forceful attention or focus.  The theories accepting this separation are also known as dual-system or dual-process theories of human cognition.
In this paper, this model is referred as a Kahneman’s two-systems model of thinking because non-psychologists \cite{murdock2012kahneman,kannengiesser2019design,preisz2019fast}] have started using it to understand the cognitive processes involved in their domains after the publication of the book titled {\em Thinking, Fast and Slow} \cite{kahneman2011thinking}.

Kahneman's model primarily consists of two systems, as shown in Figure \ref{2systems}.  These systems are labelled as {\em System 1} and {\em System 2}.  These systems can be considered as mental constructs or fictitious agents driving the thought process.  
The {\em System 1} usually responds to routine operations quickly in an unconscious way while {\em System 2} is called in action in a novel situation, responds consciously and slowly. In comparison with {\em System 2}, the {\em System 1} is more error-prone and unreliable. The following examples illustrate the existence of such two different modes of thinking in the context of Basic Algebra.

\subsubsection*{Example 1} We answer the question {\em 2 + 2 = ?} without any effort, quickly, and accurately. But, to answer the question {\em 17 X 24 =?}, we require to put effort and do deliberate calculations.

\subsubsection*{Example 2} Consider the following example from Kahneman's book \cite{kahneman2011thinking} page 44:
\begin{quote}
    A bat and ball cost \$1.10.
The bat costs one dollar more than the ball.
How much does the ball cost?
\end{quote}
Most of the people answer the question as  $10 \cent$ using their intuitive thinking ({\em System 1})  which is a wrong answer, while the correct answer is $5 \cent$.

This example is purposefully designed to demonstrate that {\em System 1} is error-prone.

\begin{table}[t]
    \centering
    \caption{Program 1 and Program 2}
    \label{programs}
    \begin{tabular}{|p{2.25in}|p{2.25in}|}
    \hline
        Program 1 &  
        Program 2   \\ \hline 
    
        \begin{lstlisting}
int main() 
{ 
int x=10, y=20; 
    
int temp = x; 
x = y; 
y = temp; 
  
printf("x= %d,y= %d", x,y); 
return 0; 
}    
\end{lstlisting} &

\begin{lstlisting}
int main() 
{ 
int x = 10, y = 20; 

x = x + y;
y = x - y; 
x = x - y; 

printf("x= %d,y= %d", x,y);
return 0;
} 
\end{lstlisting}

\\ \hline 
    \end{tabular}
    \end{table}
    
\subsection*{Example 3} Let us consider the third example. This is a question asked in the end-semester examinations of First course on Computer Programming (C Programming) offered at Dr. B. A. Tech. University India to its undergraduate students of Engineering. 

\begin{quote}
    {\bf Question:} Which of the program(s) shown in Table \ref{programs} swaps the values  of the variables $x$ and $y$?
    \begin{enumerate}
        \item [(A)] Program 1
       \item [(B)]  Program 2
    \item[(C)] None of the programs Program 1 or Program 2.
    \item [(D)] Both Program 1 and Program 2.
    \end{enumerate}
 \end{quote}
While answering this question, majority of the students (77\%)  out of the eighty three students enrolled for the course have answered  it as $(A)$ i.e. Program 1 while the correct answer is $(D)$ i.e. Both Program 1 and Program 2. This example demonstrates that, most of the students relied on their {\em System 1} while answering the question.

\section{Characteristics of Dual System Thinking }

The examples in the previous section demonstrate that thinking happens in two different modes.  To elaborate the working further, we describe some of the characteristics crucial for understanding computational thinking. This section describes these characteristics in a general setting. However, these characteristics also apply in the context of computational thinking in its broader sense when we presume human as a computing agent. 
\begin{enumerate}
    \item {\bf Both systems work concurrently} The {\em System 1 } and {\em System 2 }  work concurrently and cooperate most of the times while thinking and reasoning about the external world. When conflict arises, {\em System 2} attempts to regulate {\em System 1}. The {\em System 1} generates feelings, impressions, and inclinations. When asked by {\em System 1}, the {\em System 2} endorses and transforms them into beliefs and attitudes. Neither system can be turned off but {\em System 2} is lazy to respond as compared to {\em System 1 }.
    
    \item {\bf  The two-systems model represents a division of labour} The {\em System 1} performs some tasks efficiently while others are performed by {\em System 2}. The {\em System 1} performs tasks such as: (i) To execute skilled responses after imparting proper training. For example, applying brakes while driving. (ii) Recognizing typical situations, recognizing norms and standards and complying to conventions. For instance, recognizing irritations in the voice of a known person.  (iii) To identify causes and intentions. For example, identifying reasons behind delayed arrival of an aircraft.    
    
    The {\em System 2} is called in action when a task requires more effort and attention, such as filling a form to apply for graduate studies or selecting a  University for graduate studies. 
    
    \item{\bf Biases and heuristics guide the {\em System 1} responses} The responses of {\em System 1} are quick and  error-prone. They are quick because heuristics drives them and they are error-prone because biases guide them.  For example, we often quickly judge the level of confidence of a person through the external attributes such as being well-dressed and well-groomed, which is an instance of the use of a heuristic called {\em halo effect}. For example, the {Example 3} from the previous section,  students who responded with the wrong option $(A)$, they relied on a bias called {\em availability bias}. The {\em availability bias} selects a familiar and widely exposed option over the least exposed and un-familiar one. The option $(A)$ fulfills this criteria and majority of the students select it.
\end{enumerate}
These characteristics play a significant role in understanding human thought process in general and computational thinking in our case.

\begin{table}[t]
    \centering
    \caption{Computational Thinking Activities}
    \label{p2}
    \begin{tabular}{|p{3in}|p{3in}|}
    \hline
        {\em System 1} &  
        {\em System 2}   \\ \hline 
    \begin{enumerate}
      \item  Deciding upon primitive data types for variables.
      \item  Separating user-defined identifiers from language defined one.
      \item Identifying input and outputs of a given function or Application Programming Interface.
      \item Separating comments from code segment.
      \item Performing simple operations  like $push$, $pop$, $front$, and $rear$ on a given data structure such as {\em stack} and {\em queue}.
      \item Finding the root and leaf nodes of a given binary tree.
      \item Categorizing machine learning problem as a linear regression, classification or logistic regression, or a clustering problem from a given scatter plot.
  \end{enumerate}
   &
   \begin{enumerate}
      \item  Decomposing a large program into smaller reusable  programs. 
      \item Deciding upon a data structure (e.g., Stack, Queue, Tree, Graph) to realize the solution for a given problem.
      \item Performing tree or graph traversals (e.g., Pre-order, Post-order, Breadth-First, and Depth-First).
      \item Deciding upon  when to use a data-driven or Machine Learning approach and an algorithmic approach.
      \item Separating concerns such as business logic from communication, coordination and other such concerns.
      \item Answering the question:{\em how difficult is it to solve a give problem through computer?} or performing complexity analysis.
      \item  Recognizing the situations where approximate solution may be sufficient. 
      \item Code inspection for compliance and violation of  programming guidelines such as names of classes in Object Oriented programs shall be noun and the name of a method shall be verb.
  \end{enumerate}

\\ \hline 
    \end{tabular}
    \end{table}

\section{ Two-Systems Model and Computational Thinking}

We often consider computational thinking is a deliberate thought process driven by the goals to be achieved. So it is a slow and effortful activity requiring a high level of focus and attention. Hence, we may conclude that computational thinking is a domain of {\em System 2}, and there is no or minimal role to play for {\em System 1}. In this section, we hypothesize that two systems govern computational thinking. First one is fast, automatic and intuitional. The second one is slow, effortful,
and systematic. 

To support our argument, we identify smaller and primitive computational thinking activities and map them on {\em System 1} and {\em System 2}.  Table \ref{p2} shows some of the computational thinking activities mapped to {\em System 1} and {\em System 2}. 

While defining this mapping, we assumed a minimum level of knowledge and skills that students acquire after the courses on Computer Programming, Data Structures and Algorithm, Software Engineering, and Theory of Computation. This requirement is typically satisfied by the students studying in the final year of Computer Science and Engineering programs offered at Indian Universities. 

The mapping in the  Table \ref{p2} is based on random analyses of students' responses to the questions asked from different examinations of courses on Computer Programming, Data Structures and Algorithms, Software Engineering, and Theory of Computation.

However, the mapping can be validated by conducting intentional examinations to observe students response time and other physiological parameters such as dilation of pupil, blood pressure and heart rate.  Researchers from Psychology found that when someone is engaged in System 2 thinking activities, heart rate increases, blood pressure raises, and eye pupils dilate. 

The list in Table \ref{p2} is a suggestive and not comprehensive one which can be extended by including activities from Software Design, Project Management and other higher-level cognitive activities.

\section{Applications of Dual System Model in  Computer Science}
Many Computer Science Researchers have started taking an interest in dual-system theory from the field of Psychology and applying it in various ways. This section briefly reviews some of the recent approaches to place our approach in the proper context. These approaches can be broadly classified into three categories.
\begin{enumerate}
    \item {\bf To decompose information processing systems} These approaches assume that information processing is a complex activity which can be decomposed into two parts. The first one that requires fast responses with an acceptable level of accuracy and the second one is requiring slow but correct reasoning.  A complex information task is then divided into these lines.
    
    For example, Di Chen et al. \cite{chen2019deep}, develop a two-systems approach to solve Sudoku puzzles and the Crystal Structure Phase mapping problem. They use deep neural networks to design {\em System 1 }, which performs the tasks of pattern recognition and heuristic evaluation. The responsibility of constraint reasoning is delegated to {\em System 2 } which adopts Integer programming and Monte Carlo Tree Search techniques for this purpose. 
    
    In another example,  Sudeep Mittal et al. \cite{mittal2017thinking}
   adopt a similar two-systems approach to represent knowledge and reasoning. They use the structure called Vector Space, as a  fast-thinking processor,  to recognize patterns. Further, they use the structure called Knowledge Graph, as a slow thinking processor to reason about complex dependency relations.
   
   Some of the other similar approaches include decomposing the optimization problem at a global level in Smart Grid system  \cite{goel2017thinking} and to build ethical AI applications \cite{rossi2019preferences}. 
    
     \item {\bf To analyze the role of Biases in Software Engineering} The Kahneman's model of thinking is a rich framework for cognitive analysis, and it has been found useful to investigate human aspects of various software engineering and project management activities.  Some of these studies, \cite{mohanani2018cognitive,ruping2014taming,behimehr2020cognitive,ccalikli2013influence,zalewski2020cognitive}, have analyzed the effect of cognitive biases and heuristics on software engineering and project management activities.
    
    \item {\bf To analyze cognitive activities}  The researchers have been using the two-systems model of thinking to explain the higher-level cognitive tasks.  For example, Maria Csernoch \cite{csernoch2017thinking} use Kahneman's model to validate the observations in the study conducted to analyze errors in Computer Problem Solving by non-professionals. In the Second example, Udo Kannengiesser et al. \cite{kannengiesser2019design} analyze the process of design thinking to decompose it into smaller activities and map the lower activities as fast and slow thinking activities.
   
\end{enumerate}
The work presented in this paper breaks down the high-level Computational Thinking task to smaller activities and map them on two systems as done by Udo Kannengiesser et al. \cite{kannengiesser2019design} for the task of design thinking.
\section{Conclusion}

The paper identifies the necessity of investigating the psychological dimension of Computational Thinking skill. Further, it proposes to adopt Kahneman's Two-systems model of thinking for this purpose because it is simple to utilise, and it is rich enough in terms of analytical tools. Primarily, it separates the human thought process in two broad categories: (i) Fast and intuitional activities, and (ii) Slow and deliberate one.  

The paper illustrates the applicability of the approach by mapping CT activities on two systems requiring fast and slow thinking as a baseline for further empirical investigation.   The identified mapping needs to be substantiated by carrying out either controlled experiments or through the detailed analyses of students' responses in an educational setting.

Kahneman's two-systems model is rich as a cognitive analysis framework providing a broad set of biases and heuristics, which can be used to study the human aspects of computational thinking. It will be interesting to explore the role of these biases and heuristics in the context of Computational Thinking to make it less error-prone and a faster reasoning activity. 

\section*{Acknowledgement} The first author would like to acknowledge his younger brother Mr. Nagesh Kiwelekar for inspiring him to explore the connections between dual system theories  and Computer Science.
\bibliographystyle{plain}
\bibliography{paper}

\end{document}